\documentclass[runningheads]{llncs}

\usepackage[utf8]{inputenc}

\usepackage{graphicx}

\usepackage{tikz}
\usepackage{pgfplots}

\usepackage{amssymb}
\usepackage{amsmath}

\usepackage{xargs}      % Use more than one optional parameter in a new commands. Verwenden wir in todonotes
\usepackage{todonotes}

\newcommandx{\flo}[2][1=]{\todo[linecolor=green,backgroundcolor=green!20,bordercolor=green,#1]{\normalsize{#2}}}
\newcommandx{\floin}[2][1=]{\todo[inline,linecolor=green,backgroundcolor=green!20,bordercolor=green,#1]{\normalsize{#2}}}

\usepackage[boxed]{algorithm2e}
\usepackage{hyperref}
\hypersetup{colorlinks=true,urlcolor=blue,citecolor=blue,linkcolor=red}
\usepackage{floatrow}

\usepackage{siunitx}
\usepackage{booktabs}
\usepackage{float}

\newcommand{\bftab}{\fontseries{b}\selectfont}

\newcommand{\ie}{i.\,e.,\ }
\newcommand{\eg}{e.\,g.,\ }
\newcommand{\Eg}{E.\,g.,\ }

\newcommand{\wrt}{w.\,r.\,t.\ }
\newcommand{\etal}{et al.\ }

\newcommand{\pa}{par2}
\newcommand{\score}{score}
\newcommand{\bwuni}{bwUniCluster}

\newcommand{\introduceterm}[1]{{\emph{#1}}}

\newcommand{\refsec}[1]{Section~\ref{#1}}

\newcommand{\Reftwosecs}[2]{Sections~\ref{#1} and~\ref{#2}}
\newcommand{\figuretext}{Fig.}
\newcommand{\reffig}[1]{\figuretext~\ref{#1}} % possibility to write in small letters
\newcommand{\Reffig}[1]{\figuretext~\ref{#1}} % two commands, in case of sentence beginning

\newcommand{\reftab}[1]{Table~\ref{#1}}
\newcommand{\Reftab}[1]{Table~\ref{#1}}

\newcommand{\refalg}[1]{Algorithm~\ref{#1}}

\newcommand{\algoformat}[1]{\textsc{#1}}

\newcommand{\OurSolver}{\algoformat{GapSAT}} % Glucose assisted probSAT
\newcommand{\sr}{\algoformat{Sparrow2Riss}}
\newcommand{\prob}{\algoformat{probSAT}}
\newcommand{\glu}{\algoformat{Glucose}}

\newcommand{\setsofvarsorlit}[2]%
{\mathrm{#1}({#2})}

\newcommand{\vars}[1]{\setsofvarsorlit{Vars}{#1}}

\newcommand{\tvastd}{{\ensuremath{\alpha}}}

\begin{document}
\title{On the Effect of Learned Clauses on Stochastic Local Search\thanks{The authors acknowledge support by the state of Baden-Württemberg through bwHPC.}}
\author{Jan-Hendrik Lorenz \and Florian Wörz\thanks{Supported by the Deutsche Forschungsgemeinschaft (DFG).}}
\authorrunning{J. Lorenz \and F. Wörz}
\institute{Ulm University, Institute of Theoretical Computer Science, 89069 Ulm, Germany 
	\email{\{jan-hendrik.lorenz,florian.woerz\}@uni-ulm.de}\\
}
\maketitle              % typeset the header of the contribution
\begin{abstract}
	There are two competing paradigms in successful SAT solvers: Conflict-driven clause learning (CDCL) and stochastic local search (SLS).
	CDCL uses systematic exploration of the search space and has the ability to learn new clauses.
	SLS examines the neighborhood of the current complete assignment.
	Unlike CDCL, it lacks the ability to learn from its mistakes. 	
	This work revolves around the question whether it is beneficial for SLS to add new clauses to the original formula.
	We experimentally demonstrate that clauses with a large number of correct literals \wrt a~fixed solution are beneficial to the runtime of SLS.
	We call such clauses \introduceterm{high-quality} clauses.
	
	Empirical evaluations show that short clauses learned by CDCL possess the high-quality attribute.
	We study several domains of randomly generated instances and deduce the most beneficial strategies 
	to add high-quality clauses as a preprocessing step.
	The strategies are implemented in an SLS solver, and it is shown that this considerably improves the state-of-the-art 
	on randomly generated instances. The results are statistically significant.

\keywords{Stochastic Local Search \and Conflict-Driven Clause Learning \and Learned Clauses.}
\end{abstract}

\section{Introduction}

The \introduceterm{satisfiability problem} (\introduceterm{SAT}) asks to determine if a given propositional formula $F$ has a satisfying assignment or not.
Since Cook's NP-completeness proof of the problem~\cite{Cook71ComplexityTheoremProving}, SAT is believed to be computationally intractable in the worst case.
However, in the field of applied SAT solving, there were enormous improvements in the performance of SAT solvers in the last 20 years. 
Motivated by these significant improvements, SAT solvers have been applied to an increasing number of areas, including bounded model checking~\cite{DBLP:journals/ac/BiereCCSZ03,DBLP:journals/fmsd/ClarkeBRZ01}, cryptology~\cite{DBLP:conf/sat/EibachPV08}, or even bioinformatics~\cite{DBLP:conf/sat/LynceM06}, to name just a few.
Two algorithmic paradigms turned out to be especially promising to construct solvers.
The first to mention is \introduceterm{conflict-driven clause learning}~(\introduceterm{CDCL})~\cite{BS97UsingCSP,MS96Grasp,MMZZM01Engineering}.
%, which can be seen as an improvement of the original DPLL algorithm~\cite{DLL62MachineProgram,DP60ComputingProcedure}.
The CDCL procedure systematically explores the search space of all possible assignments for the formula~$F$ in question, by constructing partial assignments in an exhaustive branching and backtracking search.
Whenever a conflict occurs, a reason (a conflict clause) is learned and added to the original clause set~\cite{BHMW09HandbookOfSAT,ST13SATProblem}.
The other successful paradigm is \introduceterm{stochastic local search} (see~\cite[Chapter~6]{BHMW09HandbookOfSAT} for an overview). Starting from a complete initial assignment for the formula, the neighborhood of this assignment is explored by moving from assignment to assignment in the space of solution candidates while trying to 
%maximize the number of satisfied clauses 
minimize the number of unsatisfied clauses
by the assignment (or some other criterion).
This movement is usually performed by \introduceterm{flipping} one variable of such a complete assignment. Both paradigms are described in \refsec{sec:Preliminaries} in more detail.

Besides the difference in the completeness of assignments considered during a run of those algorithms, another major difference between both paradigms is the completeness and incompleteness of the solvers (\ie being able to certify both satisfiability and unsatisfiability, or not)~\cite{ST13SATProblem}. CDCL solvers are complete, while SLS algorithms are incomplete. More interestingly for practitioners, perhaps, is the complimentary behavior these paradigms exhibit in terms of performance: CDCL seems well suited for application instances, whereas SLS excels on random formulas\footnote{It is, however, noteworthy that the winning solver in the random track of the SAT Competition 2018 was \sr{}, a CDCL solver.}. An interesting question thus is, if it is possible to combine the strength of both solvers or to eliminate the weaknesses of one paradigm by an oracle-assistance of the other. This challenge was posed by Selman et al. in~\cite[Challenge 7]{SKM97TenChallenges} (and again later in~\cite{KS03TenChallengesRedux}):
	\begin{quote}
		``Demonstrate the successful combination of stochastic search and systematic search techniques, by the creation of a new algorithm that outperforms the best previous examples of both approaches.''
	\end{quote}
This easy to state question turns out to be surprisingly challenging. There have been some advances towards this goal, as we survey below. However, the performance of most algorithms that try to combine the strength of both paradigms, so-called \introduceterm{hybrid solvers}, is far from those of CDCL solvers (or other non-hybrids), especially on application instances~\cite{ALMS10BoostingLocalSearchThanksToCDCL} or even any wider range of benchmark problems~\cite{KS03TenChallengesRedux}.
In~\cite{HLDV02AHybridApproachForSAT}, the DPLL algorithm \algoformat{Satz}~\cite{LA97HeuristicsBasedOnUnitPropagation} was used to derive implication dependencies and equivalencies between literals in %the
\algoformat{WalkSAT}~\cite{MSK97EvidenceForInvariantsInLocalSearch}.
%algorithm

The effect of a restricted form of resolution to clause weighting solvers was investigated in~\cite{APSS05OldResolutionMeetsModernSLS}. A similar approach was previously studied in~\cite{CI96AddingNewClausesForFasterLocalSearch}, where new resolvent clauses based on unsatisfied clauses at local minima and randomly selected neighboring clauses are added.

Local Search over partial assignments instead of complete assignments extended with constraint propagation was studied in~\cite{JL02LocalSearchWithContraintPropagation}. In case of a conflict, a conflict clause is learned, and local search is used to repair this conflict.
A similar approach to construct a hybrid solver using SLS as the main solver and a complete CDCL solver as a sub-solver was also studied in~\cite{AdrianBalint14PhDThesis,BHG09ANovelApproachToCombine}, where the performance of the solvers \algoformat{hybridGM}, \algoformat{hybridGP}, and \algoformat{hybridPP} was empirically analyzed.  The idea of these solvers is to build a partial assignment around one complete assignment from the search trajectory of the SLS solver. This partial assignment can then be applied to the formula resulting in a simpler one, which is solved by a complete CDCL solver.

A shared memory approach for multi-core processor architectures was proposed in~\cite{KSGS09IntegratingSystematicAndLocalSearchPradigms}. In this case, DPLL can provide guidance for an SLS solver, being run simultaneously on a different core.

The solver \algoformat{hbisat}, introduced in~\cite{FH07ANewHybridSolution}, uses the partial assignments calculated by CDCL to initialize the SLS solver. The SLS solver sends unsatisfied clauses to the CDCL solver to either identify an unsatisfiable subformula of those clauses or satisfy them. This approach was later significantly improved by Letombe and Marques-Silva in~\cite{LM08ImprovementsToHybridIncremental}.

Audemard \etal\cite{ALMS10BoostingLocalSearchThanksToCDCL} introduced \algoformat{SatHys}, where both components cooperate by alternating between them. \Eg when CDCL chooses a variable to branch, its polarity is extracted from the best complete assignment found by the SLS solver. On the other hand, CDCL helps SLS out of local minima (\ie CDCL is invoked conditionally in this solver).\\

\textbf{Our Contribution.}
Our hybrid solver \OurSolver{} differs from the approaches described above in the sense that CDCL is used as a preprocessor for \prob{}~\cite{BS12ChoosingProbabilityDistributionsSLS} (an SLS solver), terraforming the landscape in advance.
This approach eliminates, in many cases, the possibility for \prob{} to get stuck in local minima, eliminating the necessity of further, more complicated interactions between both paradigms.

We examine the question, whether it is beneficial for SLS to add new clauses to the original formula in a \introduceterm{preprocessing step}, by invoking a complete CDCL solver. As it turns out, not all additional clauses are created equal.
Experimentally, we demonstrate that adding clauses that contain a larger number of correct literals \wrt a fixed solution, drastically improves the performance of SLS solvers. However, clauses that only contain few correct literals \wrt the fixed solution can be deceptive for SLS. This effect can exponentially increase the runtime as measured in the number of flips of \prob{}, a very simple SLS solver.
In practice, one has to resort to known complete algorithms or proof systems to generate helpful clauses.
We, in particular, investigate the effect of new clauses learned by CDCL and depth-limited resolution (\refsec{sec:ComparisonToResolution}).
With the help of experiments, we conclude that CDCL (or resolution limited to depth~2) produces distinctively more helpful clauses for \prob{} than resolution limited to depth~1, as was studied in the past~\cite{APSS05OldResolutionMeetsModernSLS}. We, therefore, focus our effort on CDCL as a clause learning mechanism for \prob{}.

Motivated by these insights, we 
%These insights that were gathered during these comparisions, motivate to
% study 
study the quality CDCL-learned clauses have for SLS in more detail. In training experiments that are described in \refsec{sec:InitialTuningExperiments}, we systematically deduce parameter settings by statistical analysis that increase this quality. For example, shorter clauses learned by CDCL are more beneficial.
As it turns out, however, the specific width depends on the underlying formula.
Another interesting observation is that the amount of added clauses has to be carefully restricted.
Again, the specific restriction to use depends on the underlying formula.

To finally test the concrete effect of these ideas, we compared the performance of a newly designed solver with the winner of the 2018 random track competition, \sr{} \cite{s2r2018}. Our observations were implemented in \OurSolver{}, which forms a combination of \glu{} and \prob{}.
A comprehensive experimental evaluation on 255 instances provides statistical evidence that the performance of our proposed solver \OurSolver{} exceeds \sr{}' substantially. In particular, \OurSolver{} was able to solve more instances in just \SI{30}{\second} than \sr{} in \SI{5000}{\second}.
We present a summary of our experimental evaluation in \refsec{sec:ExperimentalEvaluation}.

\section{Preliminaries}
\label{sec:Preliminaries}

We briefly reiterate the notions necessary for this work. For a thorough introduction to the 
field, we refer the reader to \cite{ST13SATProblem}.
A \introduceterm{literal} over a Boolean variable $x$ is either $x$ itself or its negation $\overline{x}$.
A \introduceterm{clause} $C = a_1 \lor \dots \lor a_\ell$ is a (possibly empty) disjunction of literals $a_i$ over pairwise disjoint variables.
%We let~$\emptycl$ denote the contradictory \introduceterm{empty clause} (the clause without any literals).
A \introduceterm{CNF formula} $F = C_1 \land \dots \land C_m$ is a conjunction of clauses.
%Without loss of generality we will assume that all clauses are non-trivial in the sense that they do not contain both a literal and its negation.
A CNF formula is a \introduceterm{$k$-CNF} if all clauses in it have at~most~$k$~variables.
An \introduceterm{assignment}~$\tvastd$ for a CNF formula~$F$ is a function that maps some subset of $\vars{F}$ to $\{0,1\}$. 
Given a complete assignment~$\tvastd$, the act of changing the truth value of precisely one variable of~$\tvastd$ is called a \introduceterm{flip}.
%We denote the \introduceterm{empty assignment} with $\varnothing$.
%The standard definition of a \introduceterm{resolution derivation} of a clause $D$ from a CNF formula $F$ (denoted by $\derivof{\proofstd}{F}{D}$) is an ordered sequence of clauses $\proofstd = (C_1,\dots,C_\resendtime)$ such that $C_\resendtime = D$, and each clause $C_\restimestep$, for $\restimestep \in \nat{\resendtime}$, is either an \introduceterm{axiom clause} $C_\restimestep \in F$ or is derived from clauses $C_j$ and $C_k$ with $j,k<\restimestep$ by the \introduceterm{resolution rule}
\introduceterm{Resolution} is the proof system with the single derivation rule $\frac{B \lor x \quad C \lor \overline{x}}{B \lor C}$,
%$
%%\label{eq:ResolutionRule}
%\AxiomC{$B \lor x$}
%\AxiomC{$C \lor \overline{x}$}
%\BinaryInfC{$B \lor C$}
%\DisplayProof
%\text{\,.}
%$
where $B$ and $C$ are clauses.\\
%In the resolution rule~\eqref{eq:ResolutionRule}, we call $B \lor x$ and $C \lor \overline{x}$ the \introduceterm{parents} and $B \lor C$ the \introduceterm{resolvent}. A derivation $\refof{\proofstd}{F}$ of the empty clause from an unsatisfiable CNF formula $F$ is called \introduceterm{refutation}. Note, that resolution is a sound and complete proof system for unsatisfiable formulas in CNF.

%\todo{Introduce backbone and flip concept}

%\subsection{CDCL}

%\todo[inline]{CDCL}
\textbf{CDCL.}
CDCL solvers, introduced in~\cite{BS97UsingCSP,MS96Grasp,MMZZM01Engineering}, construct 
a partial assignment.
When some clause is falsified by the constructed assignment, the CDCL solver adds a new clause to the
original formula~$F$.
This clause is a logical consequence of~$F$.
%The basic idea behind CDCL is sketched in \refalg{alg:CDCL}.
A more detailed 
%approach
description of CDCL can be found in~\cite{BHMW09HandbookOfSAT,PD09OnThePower,ST13SATProblem}. Modern SAT solvers are additionally equipped with incremental data structures, restart policies~\cite{GSK98BoostingCominatorialSearch}, and activity-based variable selection heuristics (VSIDS)~\cite{MMZZM01Engineering}. In this work, we use the CDCL solver \glu{}~\cite{AS09Glucose} (based on \algoformat{\mbox{Mini}Sat}~\cite{ES03MiniSat}).\\

\textbf{
	%A Simple SLS Solver: 
	\prob{}.}
Contrary to CDCL-like algorithms, algorithms based on \introduceterm{stochastic local search} (\introduceterm{SLS}) operate on complete assignments for a formula~$F$. These solvers are started with a randomly generated complete initial assignment~$\tvastd$. If~$\tvastd$ satisfies~$F$, a solution is found. Otherwise, the SLS solver tries to find a solution by repeatedly flipping the assignment of variables according to some underlying heuristic. That is, they perform a random walk over the set of complete assignments for the underlying formula.

In~\cite{BS12ChoosingProbabilityDistributionsSLS}, the \prob{} class of solvers was introduced. 
Over the last few years, \prob{}-based solvers performed excellently 
on random instances: \prob{} won the random track of the SAT competition 2013, \algoformat{dimetheus} \cite{dimetheus} in 2014 and 2016, \algoformat{YalSAT} \cite{biere2017splatz} won in 2017. Only recently, in 2018, other types of solvers significantly exceeded \prob{} based algorithms. This performance is the reason for choosing \prob{} in this study. 

The idea behind the solver is that a function $f$ is used, which gives a high probability to a variable if flipping this variable is deemed advantageous. A description of \prob{} is given in \refalg{alg:probSAT}. This class of solvers is related to Schöning's random walk algorithm introduced in~\cite{Schoening02AProbabilisticAlgorithm}.

\begin{algorithm}[htb]
	%\SetAlgoLined
	\SetAlgoVlined
	\DontPrintSemicolon
	\KwIn{Formula $F$, $maxFlips$, function $f$}
	%\KwOut{A satisfying assignment $\tvastd$ for $F$ or \texttt{UNKNOWN}}
	$\tvastd$ := randomly generated complete assignment for $F$\\
	\For{i = 1 \KwTo maxFlips}{
		\lIf{$\tvastd$ satisfies $F$}{\textbf{return} ``satisfiable''}
		Choose a clause $C = (u_1 \lor u_2 \lor \dots \lor u_\ell)$
		that is falsified under~$\tvastd$\\
		%with $C \tvastd = 0$\\
		Choose $j \in \{1,\dots, \ell\}$ with probability $\frac{f(u_j,\tvastd)}{\sum_{u \in C} f(u,\tvastd)}$\\
		Flip the assignment of the chosen variable $u_j$ and update $\tvastd$
	}
	\caption{probSAT without restarts.}
	\label{alg:probSAT}
\end{algorithm}

In~\cite{BS12ChoosingProbabilityDistributionsSLS}, the \introduceterm{break-only-poly-algorithm} with $f(x,\tvastd) := \big( \varepsilon + \operatorname{break}(x,\tvastd) \big)^{-b}$ was considered for 3-SAT, where $\operatorname{break}(x,\tvastd)$ is the number of clauses that are satisfied under~$\tvastd$ but will be falsified when the assignment of~$x$ is flipped. For $k \neq 3$, the \introduceterm{break-only-exp-algorithm} $f(x,\tvastd) := b^{-\operatorname{break}(x,\tvastd)}$ was studied. Balint and Schöning~\cite{BS12ChoosingProbabilityDistributionsSLS} found good choices for the parameters of these two functions. In this work, we have adopted these parameter settings.

\section{The Quality of Learned Clauses}
\label{sec:ComparisonToResolution}

In this section, we investigate the effect logically equivalent formulas have on the SLS solver \prob{}.
More precisely, we use a formula $F$ as a base and add a set of clauses $S=\{C_1, \dots, C_t\}$ to $F$ to obtain
a new formula $G := F \cup S$.
In general, adding new clauses to a formula $F$ does not yield a logically equivalent new formula $G$. 

Thus, we observe two artificial models related to the \introduceterm{backbone} (see, \eg \cite{kilby2005backbones}) and consider the 3-CNF case in the following. 
The backbone $\mathcal{B}(F)$ are the literals appearing in all 
satisfying assignments of $F$.
In the first model, each new clause consists of one backbone literal $x\in \mathcal{B}(F)$ and two literals $y,z$ such that their complements are backbone literals, \ie $y$ and $z$ do not occur in any solution. We call this the \introduceterm{deceptive model}.
In the second model, each new clause has one backbone literal and two randomly chosen literals. 
This is the \introduceterm{general model}.
%We call two formulas $F$ and $G$ logically equivalent iff for each assignment $\alpha$: $F(\alpha) = G(\alpha)$.

\begin{figure}[htb]
	\begin{minipage}[t]{0.47\textwidth}
		% This file was created by tikzplotlib v0.9.1.
\begin{tikzpicture}

\definecolor{color0}{rgb}{0.12156862745098,0.466666666666667,0.705882352941177}

\begin{axis}[
width=5.0in*0.5,
height=3.6in*0.5,
tick label style={font=\scriptsize},
log basis y={10},
max space between ticks=20,
tick align=inside,
tick pos=left,
x grid style={white!69.0196078431373!black},
xlabel={Number of added clauses},
%xmajorgrids,
xmin=11, xmax=209,
xtick style={color=black},
y grid style={white!69.0196078431373!black},
ylabel={Flips},
%ymajorgrids,
ymin=1184.50347527053, ymax=18550691.7022306,
ymode=log,
ytick style={color=black},
title=Deceptive model
]
\addplot [semithick, color0, mark=*, mark size=1, mark options={solid}]
table {%
20 2062.18964193292
40 1837.42722470573
60 2452.65499807276
80 4940.97663110752
100 13912.5084920256
120 44050.2847035464
140 176970.935888918
160 736829.075659088
180 2899567.12221223
200 11958764.1319963
};
\end{axis}

\end{tikzpicture}
	\end{minipage}
	\hfill
	\begin{minipage}[t]{0.47\textwidth}
		% This file was created by tikzplotlib v0.9.1.
\begin{tikzpicture}

\definecolor{color0}{rgb}{0.12156862745098,0.466666666666667,0.705882352941177}

\begin{axis}[
width=5.0in*0.5,
height=3.6in*0.5,
tick label style={font=\scriptsize},
log basis y={10},
max space between ticks=20,
tick align=inside,
ytick pos=right,
xtick pos=left,
x grid style={white!69.0196078431373!black},
xlabel={Number of added clauses},
%xmajorgrids,
xmin=11, xmax=209,
xtick style={color=black},
y grid style={white!69.0196078431373!black},
%ylabel={Flips},
%ymajorgrids,
ymin=25991.410638358, ymax=3170279.30027343,
ymode=log,
ytick style={color=black},
title=General model
]
\addplot [semithick, color0, mark=*, mark size=1, mark options={solid}]
table {%
20 2548397.53442599
40 1244156.395352
60 668458.000664001
80 353104.388932
100 204840.044394
120 123816.77217
140 85014.5314439998
160 60496.3763740001
180 43125.076718
200 32334.0569980001
};
\end{axis}

\end{tikzpicture}
	\end{minipage}
	\caption{On the left, the effect of \emph{deceptive} clauses is displayed on an instance with 100 variables and 423 clauses. 
		On the right, the effect of \emph{general} clauses is displayed on an instance with 500 variables and 2100 clauses.
		The $x$-axes denote the number of additional clauses, and the $y$-axes denote the average runtime of 100 runs of \prob{} as measured 
		in the number of flips.	Both $y$-axes are scaled logarithmically.}
	\label{fig:effect}
\end{figure}
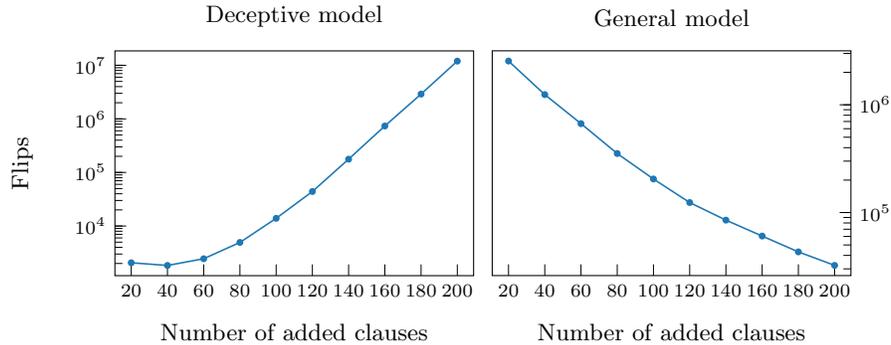

\Reffig{fig:effect} displays the effect of both models on \prob{}. On the left is the deceptive model. Generally, a large number of deceptive clauses have
a harmful effect on the runtime of \prob{}. That is, the average runtime of \prob{} increases exponentially with
the number of added clauses. 

The right-hand side of \reffig{fig:effect} shows the general model. Here, we can observe a strong, positive 
effect on the behavior of \prob{}. The average runtime of \prob{} improved by two orders of magnitude by adding 200 new clauses generated by the general model. 
Even though \reffig{fig:effect} depicts the data of only one instance, the general shape of the plot is similar on all tested instances.

Clauses generated by the deceptive model seem to give rise to new local minima which are far away from the solutions.
Once \prob{} is stuck in such a local minimum, the break-value makes it unlikely that \prob{} escapes the region of the local minimum.
On the other hand, the prevalence of correct literals in the general model seems to guide \prob{} towards a solution. 
Due to this interpretation, we call clauses that have a high number of correct literals \wrt a fixed solution \introduceterm{high-quality clauses}. The view that clauses with few correct literals have a detrimental effect on local search solvers is also 
supported by the literature~\cite{hirsch2000sat,ST13SATProblem}.

From the considerations described above, it should be evident that it is crucial which clauses are added to the formula.
Clearly, neither the deceptive nor the general model can be applied to real instances: The solution space would have to be known in advance to generate the clauses. In contrast, approaches like resolution and CDCL can be applied to 
real instances. All clauses which can be derived by resolution are already implied by the original formula. Accordingly,
adding such a clause to the original formula yields a logically equivalent formula. Similarly, clauses learned by 
a CDCL algorithm can be added to obtain a logically equivalent formula.

In the following, we compare two models based on resolution and one model based on CDCL. 
In particular, let $F$ be a formula and let $B, C \in F$
be clauses such that there is a resolvent $R$. We call $R$ a \introduceterm{level 1} resolvent. 
Secondly, let $D, E$ be clauses such that there is a resolvent $S$ and let $D$ or $E$ (or both) be level 1 resolvents.
We call $S$ a \introduceterm{level 2} resolvent.
As a representative for CDCL solvers, we use \glu{}~\cite{AS09Glucose}.

Let $F$ be a 3-CNF formula with $m$ clauses. New and logically equivalent formulas $F_1$, $F_2$, and $F_C$ are obtained in
the following manner.

\begin{center}
\begin{tabular}{l l}
	$F_1$ & Randomly select at most $m/10$ level 1 resolvents of maximum width 4  \\ & and add them to $F$.\\
	$F_2$ & Randomly select at most $m/10$ level 2 resolvents of maximum width 4 \\ & and add them to $F$. \\
	$F_C$ & Randomly select at most $m/10$ learned clauses with maximum width 4 \\ & from \glu{} (with a time limit of 300 seconds) and add them to $F$. 
\end{tabular}
\end{center}

%\noindent
The average behavior of \prob{} over 1000 runs per instance on the instance types $F_1$, $F_2$, and $F_C$ is observed.
We use a small testbed of 23 uniformly generated 3-CNF instances with \num{5000} to \num{11600} variables and a 
clause-to-variable ratio of 4.267. The instances of type $F_1$ were the most challenging for \prob{}; as a matter of fact, 
$F_1$ instances were considerably harder to solve than the original instances. On instances of type $F_2$, \prob{}
performed better, and on $F_C$, it was even more efficient. The t-test~\cite{student1908probable} confirms the observations:
$F_2$ instances are easier on average than $F_1$ instances ($p<0.01$), and $F_C$ instances are easier than $F_2$ instances
($p<0.05$). \Reftwosecs{sec:InitialTuningExperiments}{sec:ExperimentalEvaluation} present an in-depth examination of the 
effect clauses of type $F_C$ have on \prob{}.

These results lead us to believe that level 1 clauses are of low quality while level 2 and CDCL clauses are generally of higher quality. It is impractical to confirm this suspicion on uniformly generated instances of the above-mentioned size.
Hence, we use randomly generated models with hidden solutions~\cite{balyo2018using} and judge the quality of 
learned clauses based on the hidden solution. 

The SAT competition 2018 incorporated three types of models with hidden solutions. 
All three types are generated in a similar manner; they just differ in the choice of the parameters.
Here, we compare the average quality of the new clauses on each of the three models. 
For each instance, the set of all level 1, level 2, and CDCL clauses is computed,
and the quality is measured \wrt the hidden solution.

For the most part, the results confirm the observations from the uniformly generated instances:
On all three models, level 2 clauses have a statistically significantly higher quality than level 1 clauses (t-test, all $p < 0.01$).
On two of three domains, CDCL clauses have higher quality than level 2 clauses (\mbox{t-test}, both $p < 10^{-5}$), while 
level 2 clauses have higher quality on the remaining domain (t-test, $p<10^{-8}$).

As a side note, CDCL is capable of learning unit and binary clauses. Nevertheless, this did not influence the 
quality of the clauses in any meaningful way: In the 120 test instances, only a single binary and no unit 
clause was learned.

In conclusion, we conjecture that level 2 and CDCL clauses have higher quality than level 1 clauses. 
On the uniform random testbed, CDCL performs better than level 2 clauses; also, CDCL clauses have higher quality than level 2 clauses on two of the three hidden solution domains.

\section{Training Experiments}
\label{sec:InitialTuningExperiments}

In the previous section, we argued that adding supplementary clauses to an instance can have a positive effect 
on the behavior of \prob{}. The focus of this section lies on the question which clauses and how many should be added.

Especially for 3-SAT instances, an initial guess might be adding all clauses acquirable by so-called ternary
resolution \cite{billionnet1992efficient}. Informally speaking, ternary resolution is the restriction of the resolution rule to ternary clauses
such that the resolvent is either a binary or ternary clause. Ternary resolution is performed until
saturation.  
In \cite{balint2013boosting}, the effect of (amongst other techniques) ternary resolution on another SLS
solver, \algoformat{Sparrow} (see \cite{balint2010improving}), is observed. They empirically show that
ternary resolution has a negative effect on the performance on satisfiable hard combinatorial instances.
Anbulagan et al. \cite{APSS05OldResolutionMeetsModernSLS} study the effect of ternary resolution on
uniform random instances. They found that SLS solvers do not benefit from ternary resolution. 
They even conjecture that ternary resolution has a harmful impact on the runtime of SLS solvers on uniform
instances. 
We performed some experiments on our own and can confirm this suspicion for \prob{}. 
On medium-sized uniform instances, ternary resolution slowed \prob{} down by $0.5\%$ on average.
As a consequence, we focus on methods to improve the runtime behavior of \prob{} with clauses learned by 
\glu{} for the rest of this work.

The supplementary clauses are all learned by \glu{} within a 300 second time window; we only distinguish the learned clauses by their width.
The number of supplementary clauses is measured in percent of the number of original clauses.
To put it differently, we are interested in the maximal length of the new clauses and what percentage of the modified formula should
be new clauses. 
The results of this section are used to configure \OurSolver{}.

%%%%%%%%%%%%%%%%%%%%%%%%%%%%%%%%%%%%%%%%%%%%%%%%%%%%%%%%%%%%%%%%%%%%%%%%%%%%%%%%%%%%%%%%%%%%%%%%%%%%%%%%%%%

\subsection*{Description of Training Experiments}
We split the experiments into two phases. 
In the preliminary phase, promising intervals for the \textit{maximal width} and the \textit{maximal percentage}
of new clauses are obtained. In the subsequent phase, the most advantageous parameter combination is sought.
Hereafter, we describe the setup of the experiments and their results.\\

\textbf{Training Data.}
We used a set of training instances $\mathcal{C}$, which is assembled as follows:
%consisted of
All instances of the SAT Competitions random tracks\footnote{See~\url{http://www.satcompetition.org/}. In 2015 there was no random track.} 2014 to 2017 were gathered.
We filtered these instances by proven satisfiability: An instance was added to the training set $\mathcal{C}$ if and only if at least one participating solver showed satisfiability. 
Since not enough uniform random 3-SAT instances of medium size were in $\mathcal{C}$, we added all instances of this kind from the SAT Competition 2013 as well.
In total, $\mathcal{C}$ consists of 377 instances which can be divided into the following three domains: % instance categories.% (in the following $r$ will denote the ratio of clauses to variables in the instance):
\begin{itemize}
	\item 120 randomly generated instances with a \introduceterm{hidden} solution~\cite{balyo2018using},
	\item 149 uniformly generated random 3, 5, and 7-SAT instances of \introduceterm{medium size}.
	The clause-to-variable ratio is close to the satisfiability threshold~\cite{mertens2006threshold}.%:
	\item 108 uniform random 3, 5, and 7-SAT instances of \introduceterm{huge size}, \ie with over \num{50000} 
	variables. The clause-to-variable ratio of each instance is somewhat far from the satisfiability threshold.
\end{itemize}

\textbf{Training Setup.}
The experiments were performed on the \bwuni{} and a local server. Sputnik~\cite{VLSKK15Sputnik} helped to
parallelize the trials. The setup of the computer systems is heterogeneous. Therefore, the runtimes are not directly 
comparable to another.  Consequently, we do not use the runtimes for these experiments. 
Instead, the number of variable flips performed by \prob{} is used, which is a hardware-independent performance measure.

In this section, we use a timeout of $10^9$ flips for 3-SAT instances (5-SAT: $5 \cdot 10^8$; 7-SAT: $2.5 \cdot 10^8$).
This timeout corresponds to roughly 10 minutes runtime on medium-sized instances on our hardware. 
Each instance from $\mathcal{C}$ is run 1000 times for each parameter combination. 
The primary performance indicator in this section is the number of timeouts per instance.
Furthermore, the average \introduceterm{par2} value is sometimes used as a secondary performance indicator.
The \introduceterm{par2} value is the \emph{number of flips} if a solution was found or twice the timeout otherwise.
For the rest of this work, \prob{} refers to \prob{} version SC13\_v2~\cite{BalintImplementationOfProbSAT}.

%%%%%%%%%%%%%%%%%%%%%%%%%%%%%%%%%%%%%%%%%%%%%%%%%%%%%%%%%%%%%%%%%%%%%%%%%%%%%%%%%%%%%%%%%%%%%%%%%%%%%%%%%%%

\subsection*{Results of Training Experiments}
\label{sec:ResultsOfTrainingExperimentsAndDescriptionOfGapSAT}

We conducted a thorough statistical analysis of the data that was obtained in the training experiments
described above. We describe our findings in condensed form below. The main part of the remainder of this
 section is concerned with uniform, medium-sized 
instances. The results for uniform, huge instances, and instances with a hidden solution are briefly discussed
at the end of this section. \\

\begin{figure}[htb]
	\centering	
	\floatbox[{\capbeside\thisfloatsetup{capbesideposition={left,top},capbesidewidth=4.2cm}}]{figure}[\FBwidth]
	{\caption{In this plot, \prob{} is compared on the original \mbox{3-SAT} instances and the modified instances.
			The number of added clauses of the 3-SAT instances in \% is on the $x$-axis.
			The $y$-axis is the logarithm of \pa{}(modified)/\pa{}(orig).
			The blue line denotes a linear regression fit, and the light blue area is the 95\% confidence interval obtained by \num{1000} bootstrapping steps~\cite{givens2012computational}. }\label{fig:Scatterplot_per_clauses}}
	{
		\input{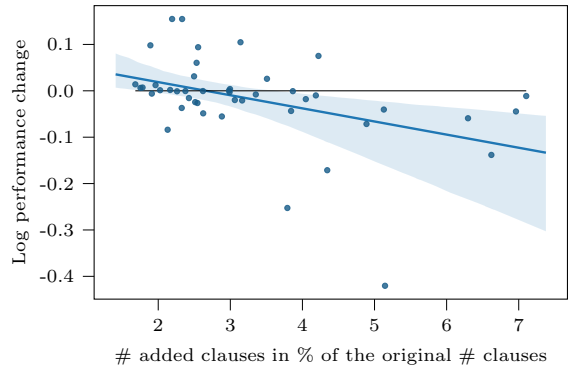}
	}
\end{figure}

\textbf{3-SAT.} 
We found that adding all clauses up to width~4 that \glu{} could find within \SI{300}{\second} 
is the most beneficial configuration. Not limiting the number of added clauses is in 
stark contrast to the 5-SAT and 7-SAT cases.
For 3-SAT, the relationship between the number of clauses and the performance is explored in \reffig{fig:Scatterplot_per_clauses}.
Each blue dot corresponds to one medium-sized 3-SAT instance from~$\mathcal{C}$. We compare the average \pa{} value on the original
instance with the average \pa{} value on the instance with all clauses up to width~4 added. Whenever the blue dot lies below the zero-baseline,
then the performance of \prob{} on the modified instance was better. The blue line is obtained by linear regression. 
By its slope, we can tell that, on average, adding more clauses is beneficial. The light blue area denotes the 95\% confidence interval,
which is calculated by bootstrapping~\cite{givens2012computational}. 
The confidence interval shows that this relationship is unlikely to be due to chance. 
We conclude that the number of new clauses should not be limited for 3-SAT instances.
On the other hand, the maximal width of the new clauses should be no more than four. 
Our experiments showed that adding longer clauses deteriorates the performance of probSAT.

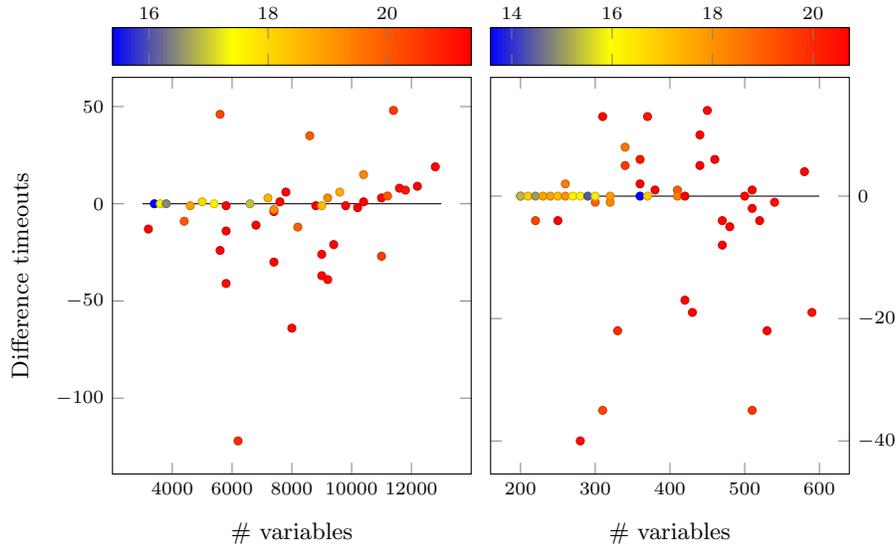
\begin{figure}[htb]
	\begin{minipage}[t]{0.47\textwidth}
		%\resizebox{.995\linewidth}{!}{
			\begin{tikzpicture}

\begin{axis}[
tick label style={font=\scriptsize},
ylabel={Difference timeouts},
xlabel={\# variables},
width=5.0in*0.5,
height=3.6in*0.75,
scaled ticks=false,
tick label style={/pgf/number format/fixed} ,
enlargelimits=true,
colorbar,
colorbar horizontal,
colorbar style={
	at={(0.5,1.03)},
	anchor=south,
	xticklabel pos=upper,
},
%colorbar style={
%	at={(0,1.2)},
%	anchor=north west,
%	 yticklabel style={
%		%align=top,
%		anchor=north west,
%		/pgf/number format/.cd,
%		fixed,
%		fixed zerofill
%	}
%},
xtick={4000,6000,8000,10000,12000},
xticklabels={4000,6000,8000,10000,12000},
%y tick label style={anchor=west,xshift=0.1cm},
]
\addplot[color=black] coordinates {(3000,0) (13000,0)};
\addplot [
only marks,
scatter,
point meta={
	\thisrow{hue}
},
mark size=1.5pt,
] table {plots/k3/k3_cdcl_300s.csv};
\end{axis}
\end{tikzpicture}
		%}
	\end{minipage}
	\hfill
	\begin{minipage}[t]{0.47\textwidth}
		%\resizebox{.995\linewidth}{!}{
		\begin{tikzpicture}
\begin{axis}[
yticklabel pos=right,
tick label style={font=\scriptsize},
xlabel={\# variables},
width=5.0in*0.5,
height=3.6in*0.75,
scaled ticks=false,
tick label style={/pgf/number format/fixed} ,
enlargelimits=true,
colorbar,
colorbar horizontal,
colorbar style={
	at={(0.5,1.03)},
	anchor=south,
	xticklabel pos=upper,
},
xtick={200,300,400,500,600},
]
\addplot[color=black] coordinates {(200,0) (600,0)};
\addplot [
only marks,
scatter,
point meta={
	\thisrow{hue}
},
mark size=1.5pt,
] table {plots/k5/k5_cdcl_5per_8.csv};
\end{axis}
\end{tikzpicture}
	%}
	\end{minipage}
	\caption{Scatterplot of the number of variables of the 3-SAT (left) and 5-SAT (right) instance against difference in timeouts between \prob{} on the original instance and the instance with the modified strategy.
	The color encodes the hardness of the original instance as measured in the logarithmic average \pa{} on \num{1000} runs of \prob{} on the original instances.}
	\label{fig:Scatterplotsmerged}
\end{figure}

The left-hand side of \reffig{fig:Scatterplotsmerged} shows the overall performance of \prob{} on instances 
with all clauses up to width~4. The $y$-axis denotes the difference of timeouts between the modified instance with all 
width~4 clauses and the original instance. Whenever the dot lies below the zero-baseline, the performance of \prob{}
was better on the modified instance. The color of the dots stands for the hardness as measured in
the average \pa{} on the original instance. We can see that adding additional clauses 
has a positive effect, especially on hard instances with 4000 to 9000 variables. Nonetheless, 
the effect reverses for more than 9000 variables. Overall the results on 3-SAT instances are not
statistically significant (t-test, $p=0.0595$). However, we believe that the main reason for this is the bad performance 
of \prob{} on the modified instances with more than 9000 variables. Furthermore, with a slightly larger sample size, the results might turn out to be statistically significant. 
We have used these observations in the configuration of \OurSolver{}, as depicted in~\reffig{fig:gapsat-flowchart}.\\

\textbf{5-SAT.} 
In preliminary experiments, we found that the maximal width of the new clauses should be in the interval $\{7,8,9\}$, and
the maximal number of new clauses should be at most $15\%$ of the original clauses. 
The effect of adding more clauses is especially pronounced: Adding more than $15\%$ of the clauses  
diminished the performance of \prob{} dramatically, in contrast to the 3-SAT case where more clauses turned out to be beneficial.

In the detailed phase of the experiments, we found that the best configuration
is adding clauses up to width~8 and using a limit of at most $5\%$ of the original clauses.
The results of this parameter configuration are shown on the right-hand side of \reffig{fig:Scatterplotsmerged}.
Again, the performance of \prob{} was better on the modified instances if the dot lies below the zero-baseline.
The color of the dots describes the hardness of the instance. Overall, the modification has a favorable impact on the performance of
\prob{} over the full domain. The effect is statistically significant (t-test, $p=0.0348$).
Also, it appears to be increasing as the number of variables increase. However, we did not further investigate this relationship.\\

\textbf{7-SAT.} 
The preliminary experiments showed that the maximal width of the new clauses should be in the interval $\{9,10,11\}$.
Moreover, similarly to the 5-SAT case, the number of new clauses should be limited. In the preliminary phase, we found
that at most $3\%$ of the original clauses should be added, otherwise the performance of \prob{} decreases.

\begin{figure}[htb]
	\centering
	\floatbox[{\capbeside\thisfloatsetup{capbesideposition={left,top},capbesidewidth=4.2cm}}]{figure}[\FBwidth]
	{\caption{Scatterplot of the number of variables of the 7-SAT instances against difference in timeouts between \prob{} on the original instance and the instance with modified strategy.
			The color encodes the hardness of the original instance as measured in the logarithmic average \pa{} on \num{1000} runs of \prob{} on the original instances. }\label{fig:Scatterplot7SAT}}
	{
		\begin{tikzpicture}
\begin{axis}[
tick label style={font=\scriptsize},
ylabel={Difference timeouts},
y label style={at={(0.04,0.5)}},
xlabel={\# variables},
width=5.0in*0.62,
height=3.6in*0.6,
enlargelimits=true,
colorbar,
colorbar horizontal,
colorbar style={
	at={(0.5,1.03)},
	anchor=south,
	xticklabel pos=upper,
},
xtick={80,100,120,140,160},
xticklabels={80,100,120,140,160},
]
\addplot[color=black] coordinates {(80,0) (170,0)};
\addplot [
only marks,
scatter,
point meta={
	\thisrow{hue}
},
mark size=1.5pt,
] table {plots/k7/k7_cdcl_1per_9.csv};
\end{axis}
\end{tikzpicture}
	}
\end{figure}
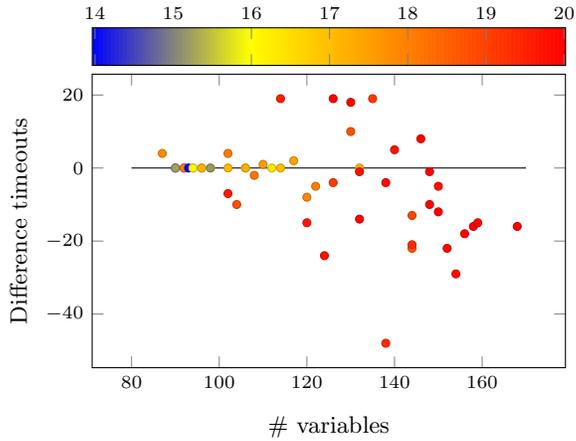

The detailed phase showed that clauses up to width~9 and a limit of at most $1\%$ is the most advantageous combination.
\reffig{fig:Scatterplot7SAT} shows the results of this combination. The modified strategy was better on average if 
the corresponding dot lies below the zero-baseline. Again, we observe that the performance of \prob{} clearly
benefits from the modified instances, especially on hard instances (red dots). 
This observation is also confirmed by the t-test ($p=0.0062$). Additionally, similar to 5-SAT instances, 
the effect seems to increase as the number of variables increase. \\

\textbf{Hidden Solution.} 
In our training set $\mathcal{C}$, all instances with a hidden solution are 3-SAT instances with few variables (at most 540).
The results are similar in nature to those discussed in the paragraph about uniform medium 3-SAT instances.
That is, the addition of new clauses of maximal width~4 and no limit on the number of new clauses is generally beneficial.\\

\textbf{Huge Instances.}
The huge instances in the training set $\mathcal{C}$ often have several million original clauses. 
In contrast, only a few new clauses are learned during the preprocessing time. Consequently, the effect of
additional clauses is negligible. The preprocessing step should, therefore, be avoided on these instances.

%%%%%%%%%%%%%%%%%%%%%%%%%%%%%%%%%%%%%%%%%%%%%%%%%%%%%%%%%%%%%%%%%%%%%%%%%%%%%%%%%%%%%%%%%%%%%%%%%%%%%%%%%%%

\subsection*{Description of \OurSolver{}}

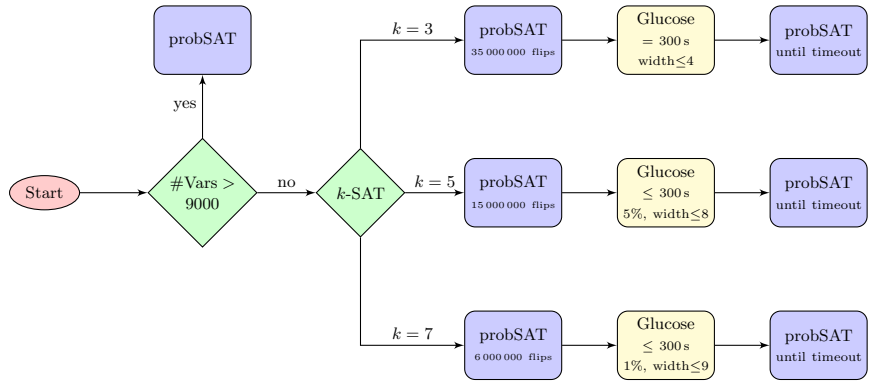
\begin{figure}[htb]
	\centering
		\usetikzlibrary{shapes,arrows}

%======================================================
% Define block styles
\tikzstyle{decision} = [diamond, draw, fill=green!20, 
text width=4.5em, text badly centered, node distance=3cm, inner sep=0pt]
\tikzstyle{block} = [rectangle, draw, fill=blue!20, 
text width=5em, text centered, rounded corners, minimum height=4em]
\tikzstyle{blockyellow} = [rectangle, draw, fill=yellow!20, 
text width=5em, text centered, rounded corners, minimum height=4em]
\tikzstyle{line} = [draw, -latex']
\tikzstyle{cloud} = [draw, ellipse,fill=red!20, node distance=3cm,
minimum height=2em]

\begin{tikzpicture}[node distance = 2.9cm,auto,font=\footnotesize, scale=0.7, transform shape]
%% Place nodes -----------------------------------------------------
\node [cloud] (start) {Start};
\node [decision, right of=start] (vars9000) {$\# \text{Vars} > 9000$};% {$\operatorname{Vars}(F) > 9000$};
\node [decision, right of=vars9000] (ksat) {$k$-SAT};
\node [block, above of=vars9000] (simpleprobsat) {probSAT};
\node [block, right of=ksat] (probsat15mio) {probSAT\\{\tiny \mbox{15\,000\,000 flips}}};
\node [block, above of=probsat15mio] (probsat35mio) {probSAT\\{\tiny \mbox{35\,000\,000 flips}}};
\node [block, below of=probsat15mio] (probsat6mio) {probSAT\\{\tiny \mbox{6\,000\,000 flips}}};
\node [blockyellow, right of=probsat35mio] (glucose3sat) {Glucose\\{\scriptsize \mbox{$=$ 300\,s}}\\{\scriptsize width$\leq$4}};
\node [blockyellow, right of=probsat15mio] (glucose5sat) {Glucose\\{\scriptsize \mbox{$\leq$ 300\,s}}\\{\scriptsize 5\%, width$\leq$8}};
\node [blockyellow, right of=probsat6mio] (glucose7sat) {Glucose\\{\scriptsize \mbox{$\leq$ 300\,s}}\\{\scriptsize 1\%, width$\leq$9}};
%
%\node [blockyellow, right of=glucose3sat] (add) {add max. 10\% of orig. clauses};
%\node [block, right of=add] (end3) {probSAT\\{\scriptsize \mbox{until timeout}}};
\node [block, right of=glucose3sat] (end3) {probSAT\\{\scriptsize \mbox{until timeout}}};
\node [block, below of=end3] (end5) {probSAT\\{\scriptsize \mbox{until timeout}}};
\node [block, below of=end5] (end7) {probSAT\\{\scriptsize \mbox{until timeout}}};
%% Draw edges -----------------------------------------------------
\path [line] (start) -- (vars9000);
\path [line] (vars9000) -- node {no} (ksat);
\path [line] (vars9000) -- node {yes} (simpleprobsat);
\path [line] (ksat) |- node [near end] {$k=3$} (probsat35mio);
\path [line] (ksat) -- node {$k=5$} (probsat15mio);
\path [line] (ksat) |- node [near end] {$k=7$} (probsat6mio);
\path [line] (probsat35mio) -- (glucose3sat);
\path [line] (probsat15mio) -- (glucose5sat);
\path [line] (probsat6mio) -- (glucose7sat);
\path [line] (glucose5sat) -- (end5);
\path [line] (glucose7sat) -- (end7);
%
%\path [line] (glucose3sat) -- (add);
%\path [line] (add) -- (end3);
\path [line] (glucose3sat) -- (end3);
\end{tikzpicture}%
	\caption{Flowchart description of GapSAT.}
	\label{fig:gapsat-flowchart}
\end{figure}

The name \OurSolver{} stands for \introduceterm{Glucose assisted probSAT}, hinting towards the combination of \prob{} as the core solver, that is being helped by a \glu{} preprocessing phase. The exact functioning principle of \OurSolver{} is depicted in the flowchart of \reffig{fig:gapsat-flowchart}.

As was noticeable in \reffig{fig:Scatterplotsmerged}, if the 3-CNF formula contained more than approx.\ \num{9000} variables, the act of adding new clauses slows down \prob{}. Furthermore, on huge instances, the preprocessing step yields no advantage.
Thus, for over \num{9000} variables, the strategy of \OurSolver{} falls back to just \prob{} on the original formula.
Otherwise, %as was mentioned in \refsubsec{subsec:RelatedWork}, 
in each case, a short run of \prob{} is used to filter out very easy to solve instances. The runtime is limited by the number of flips.
If the instance could not be solved, we employ the strategy (depending on the maximal clause width in the formula) that was deemed most promising in the evaluations described in the previous subsection. That is, we first let \glu{} extract clauses. The runtime of glucose was limited by \SI{300}{\second} in all cases. In the 5-SAT and 7-SAT case, \glu{} could finish earlier, if the restrictions on the number of added clauses were met. We again emphasize that \reffig{fig:Scatterplot_per_clauses} explains the difference between the 3-SAT case when compared to the 5-SAT and 7-SAT case. Not restricting the number of learned clauses in the 3-SAT case turned out to be the superior strategy in our training experiments.
One should further observe that \glu{} has the possibility to solve the instance during its runtime.
If this was not successful, \prob{} is restarted on the formula, that was modified by running \glu{} and adding the clauses corresponding to the strategy as developed in the previous subsection.
%We mention again, that \reffig{fig:Scatterplot_per_clauses} explains the difference of the 3-SAT case when compared to the 5-SAT and 7-SAT case. Not restricting the number of learned claueses in the 3-SAT case turned out to be the far superior strategy in our training experiments.
%
It is noteworthy that \OurSolver{} does not use any additional preprocessing techniques.
We refer to \refsec{sec:ConclusionAndOutlook} for a further discussion of that point.

\section{Experimental Evaluation}
\label{sec:ExperimentalEvaluation}

In the following, the performance of \OurSolver{} is evaluated.
We compare \OurSolver{} with the winner of the random track at the SAT competition 2018, \sr{}, 
and with the original version of \prob{}.

All experiments were executed on a computer with 32 Intel Xeon E5-2698 v3 CPUs running at \SI{2.30}{\giga\hertz}.
We set the time limit to \num{5000} seconds and used no memory limit.
The benchmarks consist of all 255 instances of the random track at the SAT competition~2018. 
Unlike the experiments in \refsec{sec:InitialTuningExperiments}, 
the performance of each solver is measured based on its \pa{} value \wrt the runtime \emph{in seconds}.
In the following, the \textit{\score{}} denotes the sum of the \pa{} values over all instances.

\begin{figure}[htb]
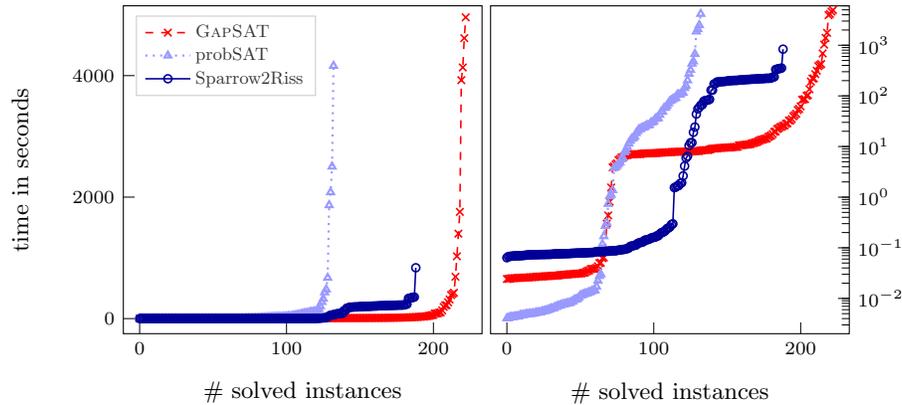

	\begin{minipage}[t]{0.47\textwidth}
		\input{plots/cactus}
	\end{minipage}
	\hfill
	\begin{minipage}[t]{0.47\textwidth}
		\input{plots/log_cactus}
	\end{minipage}
	\caption{Cactus plot comparing \prob{}, \sr{}, and \OurSolver{} on the instances of the random track of the SAT competition 2018. On the left, the plot is linearly-scaled; on the right, it is logarithmically-scaled.}
	\label{fig:cactus}
\end{figure}

\begin{table}[htbp]
	\centering
	\floatbox[{\capbeside\thisfloatsetup{capbesideposition={left,top},capbesidewidth=5.6cm}}]{table}[\FBwidth]
	{\caption{\OurSolver{}, \sr{}, and \prob{} are compared based on the number of solved instances and the corresponding \score{}. }\label{tab:comparison-all}}
	{\setlength{\tabcolsep}{5pt}
		\begin{tabular}{l | r r}
			& \# solved & \score{} \\
			\hline
			\prob{} & 133 & 1\,234\,986.01 \\
			\sr{} & 189 & 672\,335.89 \\
			\OurSolver{} & \bftab{223} & \bftab{347\,156.40} \\
		\end{tabular}
	}
\end{table}

As can be observed in~\reftab{tab:comparison-all}, \OurSolver{} solved substantially more instances than \prob{} and \sr{}.
The \score{} of \OurSolver{} is nearly halved compared to the score of \sr{}. \Reffig{fig:cactus} demonstrates that \OurSolver{}
is especially efficient within the first few seconds. \OurSolver{} solved more instances within 30 seconds
than \sr{} solved within the standard timeout of \num{5000} seconds is a case in point. This behavior can be observed in the logarithmically-scaled part of~\reffig{fig:cactus}. Furthermore, there are no instances that could be solved with \sr{}, but not with \OurSolver{} within \SI{5000}{\second}.

We used statistical testing to evaluate the performance of \OurSolver{} compared to \sr{} and \prob{}.
The t-test \cite{student1908probable} shows that the \score{} of \OurSolver{} is better than both other solvers.
We also used the Wilcoxon signed-rank test \cite{wilcoxon1945individual} to show that the median runtime of \OurSolver{} is superior
to \sr{} and \prob{}. All results are statistically significant, with $p$-values less than $10^{-9}$.
Cohen's $d$ value \cite{cohen2013statistical} is $0.39$ for the comparison with \sr{} and $0.73$ for the comparison with \prob{}.

The instances of the random track at the SAT competition 2018 can be split into three domains.
Some instances are generated uniformly at random with a medium number of variables and a clause-to-variable ratio
close to the satisfiability threshold \cite{mertens2006threshold}. Similarly, there are uniform random instances with a huge number of variables but with a clause-to-variable ratio not too close to the phase transition. Finally, there are randomly generated instances with a hidden solution \cite{balyo2018using}.
\Reftab{tab:comparison-domains} shows the performance of all three solvers on each domain. \OurSolver{} was the 
fastest solver on all three domains. It should be stated that the performances of \OurSolver{} and \prob{} are interchangeable on huge, uniform instances since the differences are just due to random noise on this domain. 
Lastly, the average time needed to learn the new clauses was 103.17 seconds.
That said, the actual time to perform the clause learning process is much shorter on most instances: The median time is just 7.89 seconds.

\begin{table}[htbp]
	\centering
	\floatbox[{\capbeside\thisfloatsetup{capbesideposition={left,top},capbesidewidth=4.1cm}}]{table}[\FBwidth]
	{\caption{\OurSolver{}, \sr{}, and \prob{} are compared on three domains based on the \score{}. }\label{tab:comparison-domains}}
	{\setlength{\tabcolsep}{4pt}
		\begin{tabular}{l | r r r}
			& hidden & medium & huge \\
			\hline
			\prob{} & 872\,938.74 & 137\,396.83 & 224\,650.43 \\
			\sr{} & 8\,589.12 & 171\,492.91 & 492\,253.86 \\
			\OurSolver{} & \bftab{851.36}& \bftab{127\,982.19} & \bftab{218\,322.85}
		\end{tabular}
	}
\end{table}

We conclude that future generations of local search solvers should incorporate some kind of clause learning mechanisms, 
for example, as a prepossessing step as used by \OurSolver{}.

\section{Conclusion and Outlook}
\label{sec:ConclusionAndOutlook}

In this work, a novel combination of CDCL as a preprocessing step and local search as the main solver is introduced.
We empirically show on several domains that short clauses learned by CDCL have a high number of correct literals \wrt 
a fixed solution. Consequently, these new clauses guide local search solvers towards a solution. 
Using this knowledge, we design a new SAT solver \OurSolver{} which uses the CDCL solver \glu{} in a preprocessing step to find new 
clauses. It then proceeds to use \prob{} on the modified formula to find a solution. We show that \OurSolver{} improves the 
state-of-the-art on randomly generated instances.

The \OurSolver{} solver can be improved even further: Besides the techniques described in this paper, no preprocessing
steps are performed. We believe that further, finely tuned preprocessing may help to increase the performance of \OurSolver{}
on instances where it struggled to find a solution.
When tuning \OurSolver{}, we used the original settings of \prob{} (\ie we use the parameters from~\cite{BalintImplementationOfProbSAT}). The only tuned parameters are the number of new clauses and their length. An interesting direction for further research is to obtain even better performance by simultaneously tuning these parameters together with the \prob{} settings.
Furthermore, we argued that the clauses which are added to the formula have a substantial effect on the performance of SLS algorithms. Even though clauses learned by \glu{} have good properties on average, it would be beneficial to devise a 
clause selection heuristic for local search algorithms. If clauses having a negative impact on local search can be avoided,
then the overall performance of solvers like \OurSolver{} should improve significantly.
Another general question that could be investigated is about the clauses from \algoformat{MiniSAT}, that is being used in \glu{}. Clauses learned by \glu{} are generated by conflict analysis but may depend on clauses
generated by the \algoformat{MiniSAT} preprocessing. It may be the case that short clauses are missed %by the technique
because they are %simply
only considered inside the solver, but never by the learning mechanism (when generated in the preprocessing step).

\mbox{~}

\textbf{Supplementary Material.} The source code of \OurSolver{} and all evaluations are available under \url{https://doi.org/10.5281/zenodo.3776052}.

 \bibliographystyle{splncs04}
 \bibliography{refArticlesAndBooksLearntClausesNoMonth}

\begin{thebibliography}{10}
\providecommand{\url}[1]{\texttt{#1}}
\providecommand{\urlprefix}{URL }
\providecommand{\doi}[1]{https://doi.org/#1}

\bibitem{APSS05OldResolutionMeetsModernSLS}
Anbulagan, A., Pham, D.N., Slaney, J.K., Sattar, A.: Old resolution meets
  modern {SLS}. In: Proceedings of the 20th National Conference on Artificial
  Intelligence and the 17th Innovative Applications of Artificial Intelligence
  Conference ({AAAI}/{IAAI}~'05). pp. 354\nobreakdash--359 (2005)

\bibitem{ALMS10BoostingLocalSearchThanksToCDCL}
Audemard, G., Lagniez, J.M., Mazure, B., Sa{\"i}s, L.: Boosting local search
  thanks to {CDCL}. In: Proceedings of the 17th International Conference on
  Logic for Programming, Artificial Intelligence and Reasoning ({LPAR}~'10).
  Lecture Notes in Computer Science, vol.~6397, pp. 474\nobreakdash--488.
  Springer (2010)

\bibitem{AS09Glucose}
Audemard, G., Simon, L.: Predicting learnt clauses quality in modern {SAT}
  solvers. In: Proceedings of the 21st International Joint Conference on
  Artificial Intelligence ({IJCAI}~'09). pp. 399\nobreakdash--404 (2009)

\bibitem{AdrianBalint14PhDThesis}
Balint, A.: Engineering stochastic local search for the satisfiability problem.
  Ph.D. thesis, University of Ulm (2014)

\bibitem{BalintImplementationOfProbSAT}
Balint, A.: Original implementation of {probSAT} (2015), available at
  \url{https://github.com/adrianopolus/probSAT}.

\bibitem{balint2010improving}
Balint, A., Fr{\"o}hlich, A.: Improving stochastic local search for sat with a
  new probability distribution. In: International Conference on Theory and
  Applications of Satisfiability Testing. pp. 10--15. Springer (2010)

\bibitem{BHG09ANovelApproachToCombine}
Balint, A., Henn, M., Gableske, O.: A novel approach to combine a {SLS-} and a
  {DPLL}-solver for the satisfiability problem. In: Proceedings of the 12th
  International Conference on Theory and Applications of Satisfiability Testing
  ({SAT}~'09). Lecture Notes in Computer Science, vol.~5584, pp.
  284\nobreakdash--297. Springer (2009)

\bibitem{balint2013boosting}
Balint, A., Manthey, N.: Boosting the performance of sls and cdcl solvers by
  preprocessor tuning. In: POS@ SAT. pp. 1--14 (2013)

\bibitem{dimetheus}
Balint, A., Manthey, N.: dimetheus. In: Proceedings of SAT Competition 2016:
  Solver and Benchmark Descriptions. vol. B-2016-1, pp. 37\nobreakdash--38.
  Department of Computer Science Series of Publications B, University of
  Helsinki (2016)

\bibitem{s2r2018}
Balint, A., Manthey, N.: {SparrowToRiss} 2018. In: Proceedings of SAT
  Competition 2018: Solver and Benchmark Descriptions. vol. B-2018-1, pp.
  38\nobreakdash--39. Department of Computer Science Series of Publications B,
  University of Helsinki (2018)

\bibitem{BS12ChoosingProbabilityDistributionsSLS}
Balint, A., Sch{\"{o}}ning, U.: Choosing probability distributions for
  stochastic local search and the role of make versus break. In: Proceedings of
  the 15th International Conference on Theory and Applications of
  Satisfiability Testing ({SAT}~'12). Lecture Notes in Computer Science,
  vol.~7317, pp. 16\nobreakdash--29. Springer (2012)

\bibitem{balyo2018using}
Balyo, T., Chrpa, L.: Using algorithm configuration tools to generate hard
  {SAT} benchmarks. In: Proceedings of the 11th Annual Symposium on
  Combinatorial Search ({SoCS}~'18). pp. 133\nobreakdash--137. {AAAI} Press
  (2018)

\bibitem{BS97UsingCSP}
{Bayardo~Jr.}, R.J., Schrag, R.: Using {CSP} look-back techniques to solve
  real-world {SAT} instances. In: Proceedings of the 14th National Conference
  on Artificial Intelligence ({AAAI~'97}). pp. 203\nobreakdash--208. {AAAI}
  Press (1997)

\bibitem{biere2017splatz}
Biere, A.: Cadical, lingeling, plingeling, treengeling, yalsat entering the sat
  competition 2017. Proceedings of SAT Competition 2017 - Solver and Benchmark
  Descriptions  \textbf{B-2017-1},  14--15 (2017)

\bibitem{DBLP:journals/ac/BiereCCSZ03}
Biere, A., Cimatti, A., Clarke, E.M., Strichman, O., Zhu, Y.: Bounded model
  checking. Advances in Computers  \textbf{58},  117\nobreakdash--148 (2003)

\bibitem{BHMW09HandbookOfSAT}
Biere, A., Heule, M., van Maaren, H., Walsh, T. (eds.): Handbook of
  Satisfiability, Frontiers in Artificial Intelligence and Applications,
  vol.~185. {IOS} Press (2009)

\bibitem{billionnet1992efficient}
Billionnet, A., Sutter, A.: An efficient algorithm for the 3-satisfiability
  problem. Operations Research Letters  \textbf{12}(1),  29--36 (1992)

\bibitem{CI96AddingNewClausesForFasterLocalSearch}
Cha, B., Iwama, K.: Adding new clauses for faster local search. In: Proceedings
  of the 13th National Conference on Artificial Intelligence and 8th Innovative
  Applications of Artificial Intelligence Conference, ({AAAI}/{IAAI}~'96). pp.
  332\nobreakdash--337 (1996)

\bibitem{DBLP:journals/fmsd/ClarkeBRZ01}
Clarke, E.M., Biere, A., Raimi, R., Zhu, Y.: Bounded model checking using
  satisfiability solving. Formal Methods in System Design  \textbf{19}(1),
  7\nobreakdash--34 (2001)

\bibitem{cohen2013statistical}
Cohen, J.: Statistical power analysis for the behavioral sciences. Routledge
  (2013)

\bibitem{Cook71ComplexityTheoremProving}
Cook, S.A.: The complexity of theorem-proving procedures. In: Proceedings of
  the 3rd Annual ACM Symposium on Theory of Computing ({STOC}~'71). pp.
  151\nobreakdash--158 (1971)

\bibitem{ES03MiniSat}
E{\'e}n, N., S{\"o}rensson, N.: An extensible {SAT}-solver. In: 6th
  International Conference on Theory and Applications of Satisfiability Testing
  ({SAT}~'03), Selected Revised Papers. Lecture Notes in Computer Science,
  vol.~2919, pp. 502\nobreakdash--518. Springer (2004)

\bibitem{DBLP:conf/sat/EibachPV08}
Eibach, T., Pilz, E., V{\"{o}}lkel, G.: Attacking bivium using {SAT} solvers.
  In: Proceedings of the 11th International Conference on Theory and
  Applications of Satisfiability Testing ({SAT}~'08). Lecture Notes in Computer
  Science, vol.~4966, pp. 63\nobreakdash--76. Springer (2008)

\bibitem{FH07ANewHybridSolution}
Fang, L., Hsiao, M.S.: A new hybrid solution to boost {SAT} solver performance.
  In: Proceedings ot the Design, Automation and Test in Europe Conference and
  Exposition ({DATE}~'07). pp. 1307--1313 (2007)

\bibitem{givens2012computational}
Givens, G.H., Hoeting, J.A.: Computational Statistics, vol.~703. John Wiley \&
  Sons (2012)

\bibitem{GSK98BoostingCominatorialSearch}
Gomes, C.P., Selman, B., Kautz, H.A.: Boosting combinatorial search through
  randomization. In: Proceedings of the 15th National Conference on Artificial
  Intelligence and 10th Innovative Applications of Artificial Intelligence
  Conference, ({AAAI}/{IAAI}~'98). pp. 431\nobreakdash--437 (1998)

\bibitem{HLDV02AHybridApproachForSAT}
Habet, D., Li, C.M., Devendeville, L., Vasquez, M.: A hybrid approach for
  {SAT}. In: Proceedings of the 8th International Conference on Principles and
  Practice of Constraint Programming ({CP}~'02). Lecture Notes in Computer
  Science, vol.~2470, pp. 172\nobreakdash--184. Springer (2002)

\bibitem{hirsch2000sat}
Hirsch, E.A.: Sat local search algorithms: worst-case study. Journal of
  Automated Reasoning  \textbf{24}(1-2),  127--143 (2000)

\bibitem{JL02LocalSearchWithContraintPropagation}
Jussien, N., Lhomme, O.: Local search with constraint propagation and
  conflict-based heuristics. Artif. Intell.  \textbf{139}(1),
  21\nobreakdash--45 (2002), preliminary version in \emph{IAAI~'00}

\bibitem{KS03TenChallengesRedux}
Kautz, H.A., Selman, B.: Ten challenges redux: Recent progress in propositional
  reasoning and search. In: Proceedings of the 9th International Conference on
  Principles and Practice of Constraint Programming ({CP}~'03). Lecture Notes
  in Computer Science, vol.~2833, pp. 1\nobreakdash--18. Springer (2003)

\bibitem{kilby2005backbones}
Kilby, P., Slaney, J., Thi{\'e}baux, S., Walsh, T., et~al.: Backbones and
  backdoors in satisfiability. In: AAAI. vol.~5, pp. 1368--1373 (2005)

\bibitem{KSGS09IntegratingSystematicAndLocalSearchPradigms}
Kroc, L., Sabharwal, A., Gomes, C.P., Selman, B.: Integrating systematic and
  local search paradigms: {A} new strategy for {MaxSAT}. In: Proceedings of the
  21st International Joint Conference on Artificial Intelligence ({IJCAI}~'09).
  pp. 544\nobreakdash--551 (2009)

\bibitem{LM08ImprovementsToHybridIncremental}
Letombe, F., Marques{-}Silva, J.: Improvements to hybrid incremental {SAT}
  algorithms. In: Proceedings of the 11th International Conference on Theory
  and Applications of Satisfiability Testing ({SAT}~'08). Lecture Notes in
  Computer Science, vol.~4996, pp. 168\nobreakdash--181. Springer (2008)

\bibitem{LA97HeuristicsBasedOnUnitPropagation}
Li, C.M., Anbulagan, A.: Heuristics based on unit propagation for
  satisfiability problems. In: Proceedings of the 15th International Joint
  Conference on Artificial Intelligence ({IJCAI}~'97). pp. 366\nobreakdash--371
  (1997)

\bibitem{DBLP:conf/sat/LynceM06}
Lynce, I., Marques{-}Silva, J.: {SAT} in bioinformatics: Making the case with
  haplotype inference. In: Proceedings of the 9th International Conference on
  Theory and Applications of Satisfiability Testing ({SAT}~'06). Lecture Notes
  in Computer Science, vol.~4121, pp. 136\nobreakdash--141. Springer (2006)

\bibitem{MS96Grasp}
{Marques-Silva}, J.P., Sakallah, K.A.: {GRASP}---a new search algorithm for
  satisfiability. In: Proceedings of the {IEEE}/{ACM} International Conference
  on Computer-Aided Design ({ICCAD}~'96). pp. 220\nobreakdash--227 (1996)

\bibitem{MSK97EvidenceForInvariantsInLocalSearch}
McAllester, D.A., Selman, B., Kautz, H.A.: Evidence for invariants in local
  search. In: Proceedings of the 14th National Conference on Artificial
  Intelligence and 9th Innovative Applications of Artificial Intelligence
  Conference ({AAAI}/{IAAI}~'97). pp. 321\nobreakdash--326 (1997)

\bibitem{mertens2006threshold}
Mertens, S., M{\'e}zard, M., Zecchina, R.: Threshold values of random {$K$-SAT}
  from the cavity method. Random Structures \& Algorithms  \textbf{28}(3),
  340\nobreakdash--373 (2006)

\bibitem{MMZZM01Engineering}
Moskewicz, M.W., Madigan, C.F., Zhao, Y., Zhang, L., Malik, S.: Chaff:
  {E}ngineering an efficient {SAT} solver. In: Proceedings of the 38th Design
  Automation Conference (DAC~'01). pp. 530\nobreakdash--535 (2001)

\bibitem{PD09OnThePower}
Pipatsrisawat, K., Darwiche, A.: On the power of clause-learning {SAT} solvers
  with restarts. In: Proceedings of the 15th International Conference on
  Principles and Practice of Constraint Programming ({CP}~'09). Lecture Notes
  in Computer Science, vol.~5732, pp. 654\nobreakdash--668. Springer (2009)

\bibitem{Schoening02AProbabilisticAlgorithm}
Sch{\"{o}}ning, U.: A probabilistic algorithm for k-sat based on limited local
  search and restart. Algorithmica  \textbf{32}(4),  615\nobreakdash--623
  (2002), preliminary version in \emph{FOCS~'99}

\bibitem{ST13SATProblem}
Sch{\"o}ning, U., Tor{\'a}n, J.: The Satisfiability Problem: Algorithms and
  Analyses, Mathematics for Applications (Mathematik f{\"{u}}r Anwendungen),
  vol.~3. Lehmanns Media (2013)

\bibitem{SKM97TenChallenges}
Selman, B., Kautz, H.A., McAllester, D.A.: Ten challenges in propositional
  reasoning and search. In: Proceedings of the 15th International Joint
  Conference on Artificial Intelligence ({IJCAI}~'97). pp. 50--54 (1997)

\bibitem{student1908probable}
Student: The probable error of a mean. Biometrika  \textbf{6}(1),
  1\nobreakdash--25 (1908)

\bibitem{VLSKK15Sputnik}
V{\"{o}}lkel, G., Lausser, L., Schmid, F., Kraus, J.M., Kestler, H.A.: Sputnik:
  \emph{ad hoc} distributed computation. Bioinformatics  \textbf{31}(8),
  1298\nobreakdash--1301 (2015)

\bibitem{wilcoxon1945individual}
Wilcoxon, F.: Individual comparisons by ranking methods. Biometrics
  \textbf{1}(6),  80\nobreakdash--83 (1945)

\end{thebibliography}

\end{document}